\documentclass[review,12pt]{elsarticle}

\usepackage{setspace} 
\usepackage{geometry}
\usepackage{algorithm} 
\usepackage{algpseudocode} 
\usepackage{graphicx}
\usepackage{caption}
\usepackage{subcaption}
\usepackage[T1]{fontenc}
\usepackage[utf8]{inputenc}
\usepackage{calc}
\usepackage{indentfirst}
\usepackage{fancyhdr}
\usepackage{graphicx,epstopdf}
\usepackage{lastpage}
\usepackage{ifthen}
\usepackage{float}
\usepackage{amsmath}
\usepackage{amssymb} % For math environment bold format
\usepackage{enumitem}
\usepackage{mathpazo}
\usepackage{booktabs} % For \toprule etc. in tables
\usepackage{titlesec}
\usepackage{etoolbox} % For \AtBeginDocument etc.
\usepackage{tabto} % To use tab for alignment on first page
\usepackage{xcolor, colortbl} % To provide color for soul (for english editing), for adding cell color of table
\usepackage{soul} % To highlight text

\usepackage{multirow}
\usepackage{microtype} % For command \textls[]{}
\usepackage{tikz} % For \foreach used for Orcid icon
\usepackage{totcount} % To enable extracting the value of the counter "page" 
\usepackage{changepage} % To adjust the width of the column for the title part and figures/tables (adjustwidth environment)
\usepackage{attrib} % For XML2PDF use \tag{} for equation
\usepackage{upgreek} % For making greek letters not italic
\usepackage{array} % For table array
\usepackage{tabularx}
\usepackage{pbox} % For biography environment
\usepackage{ragged2e} % For command \justifying
\usepackage[]{tocloft} % For dots in TOC. If subfigure package is loaded, the subfigure option needs to be added here to avoid clash: \RequirePackage[subfigure]{tocloft}
\usepackage{marginfix} % For command \clearmargin for manually moving the left column to the next page
\usepackage{enotez}
\usepackage{marginnote}
\usepackage{url}
\usepackage{mathrsfs}

\DeclareMathOperator*{\argmin}{arg\,min}
\newlength{\leftcolumnlength}
\newlength{\extralength}
\newlength{\fulllength}
\makeatletter
\def\ps@pprintTitle{%
	\let\@oddhead\@empty
	\let\@evenhead\@empty
	\def\@oddfoot{}%
	\let\@evenfoot\@oddfoot}
\makeatother

\begin{document}
	
\begin{frontmatter}
		
		
		
		\title{Model Pruning Enables Localized and Efficient Federated Learning for Yield Forecasting and Data Sharing}

		\author[aberdeen]{Andy Li\corref{corr}}
		\ead{a.li.21@abdn.ac.uk}
		\author[aberdeen,aberdeen2]{Milan Markovic}
		
		\author[aberdeen]{Peter Edwards}
		
		\author[aberdeen,aberdeen2]{Georgios Leontidis\corref{corr}}
		\ead{georgios.leontidis@abdn.ac.uk}

		%\author[inst2]{Author Two}
		%\author[inst1,inst2]{Author Three}
		
		\affiliation[aberdeen]{organization={School of Natural and Computing Sciences, University of Aberdeen},
			city={Aberdeen},
			postcode={AB24 3UE},
			country={UK}}
		
		\affiliation[aberdeen2]{organization={Interdisciplinary Centre for Data and AI, University of Aberdeen},
			city={Aberdeen},
			postcode={AB24 3FX},
			country={UK}}
		
		\cortext[corr]{Corresponding author}
		
		\begin{abstract}
			Federated Learning (FL) presents a decentralized approach to model training in the agri-food sector and offers the potential for improved machine learning performance, while ensuring the safety and privacy of individual farms or data silos. However, the conventional FL approach has two major limitations. First, the heterogeneous data on individual silos can cause the global model to perform well for some clients but not all, as the update direction on some clients may hinder others after they are aggregated. Second, it is lacking with respect to the efficiency perspective concerning communication costs during FL and large model sizes. This paper proposes a new technical solution that utilizes network pruning on client models and aggregates the pruned models. This method enables local models to be tailored to their respective data distribution and mitigate the data heterogeneity present in agri-food data. Moreover, it allows for more compact models that consume less data during transmission. We experiment with a soybean yield forecasting dataset and find that this approach can improve inference performance by 15.5\% to 20\% compared to FedAvg, while reducing local model sizes by up to 84\% and the data volume communicated between the clients and the server by 57.1\% to 64.7\%.
		\end{abstract}
		
		%\begin{graphicalabstract} https://daac.ornl.gov/cgi-bin/628
		%dsviewer.pl?ds_id=1840
		
		
		
\end{frontmatter}
	
	\section{Introduction}
	The agri-food supply chain involves the whole journey from farm to fork, including agriculture, food processing, warehousing systems, distribution and marketing. Data analytics hold the key to ensuring food security and sustainability. Machine learning has been widely adapted to provide technical solutions to analytical problems in agriculture and food sectors, such as crop yield prediction \cite{VANKLOMPENBURG2020105709, jeong2016random, onoufriou2023premonition, alhnaity2021autoencoder},  consumption demand forecasting \cite{agriculture10010021, su12166409}, crop and disease detection \cite{7891032,10.3389/fpls.2016.01419}, quality control and intelligent scheduling \cite{rong2019computer, thota2021contrastive, onoufriou2019nemesyst}, and several others. 
	
	Typically, building such statistical models requires large amounts of data collected from various sources, i.e., different farms, supply chains, and other stakeholders. However, individually, they may not have adequate data to train competent machine learning models for the tasks. While combining their data into a centralized silo may improve data quality, collecting it may be challenging due to commercially sensitive information and reputational risks \cite{DURRANT2021100493}. Federated Learning (FL) is a well-established training algorithm that addresses this by allowing a model to be trained decentrally without physically sharing the data but instead sharing the model information only \cite{DBLP:journals/corr/McMahanMRA16}. Each participating device (referred to as a client) participates in training in an isolated environment and is coordinated by the central server. As a result, FL allows models to be collaboratively trained on large datasets while preserving data privacy, and this can be leveraged within the agri-food industry when training on local dataset alone is insufficient. 
	
	However, with the progressive improvements in deep learning models, their number of parameters have increased exponentially \cite{menghani2023efficient}. Since FL requires the transmission of the model between the server and the clients in each round, the communication cost often becomes a bottleneck \cite{konevcny2016federated}. Large models also make edge device deployment challenging as they consume more memory footprint and computational power. Moreover, in real-world scenarios, clients typically hold heterogeneous data, meaning the data distribution is different and can be diverse in nature even when the data measurements are held consistent. For example, when data from hundreds of farms is used to build a model for crop yield prediction, it often comes from different regions with inconsistent readings. This could be caused by disparities in sensor deployment or other variables that could affect the yield but not reflected in the data, such as different crop genetics and soil types. Heterogeneous data can cause the model to not generalize well, especially for farms with bigger deviations from the mean distribution. 
	
	In this paper, we propose a federated learning strategy, which we name Federated Pruning (FedPruning) to address these limitations. Neural networks are typically over-parameterized and there is much redundancy \cite{denil2014predicting}. It has been consistently shown that same inference performance can be reached with only a fraction of the original model size \cite{blalock2020state}. We leverage the theoretical benefits of pruning, and remove redundant parameters from individual client models. Through our training algorithm, all client models have different connections and weight values by the end of training, and they become localized and tailored to their own data. The resulting models not only have better local inference performance compared to the traditional method, but they are also smaller in size, more energy and memory efficient. Communication is a common bottleneck in FL, and we are motivated by the usage of machine learning models in agri-food both in the traditional broadband and Internet of Things (IoT) settings where hardware capacity and internet are more limited. Reducing the number of parameters in the models can decrease the data exchange volume of the participants, resulting in lower communication costs and feasibility of edge device implementation, considering the internet and coverage may be more unreliable and limited in rural settings.  
	
	To showcase the potential benefits of FedPruning, we employ an established dataset that has previously been used for soybean yield forecasting \cite{Khaki_2020} and FL within agri-food contexts \cite{DURRANT2022106648}. We demonstrate that our method improves inference performance of local models, reduces communication cost during training and results in smaller sizes compared to the FL baseline. In summary, this work describes the following contributions:
	
	\begin{itemize}
		\item To the best of our knowledge, this is the first study to conceptualize a communication-efficient machine learning methodology that is built upon data privacy and allows for decentralised training with neural network pruning in an agri-food setting.
		\item We propose a new pruning methodology that improves inference performance and communication efficiency at the same time in a FL setup.
		\item We demonstrate the applicability of our method on a real-world dataset and show that it outperforms both centralized and FedAvg baselines, suggesting it as a viable option in decentralized agri-food settings.
	\end{itemize}
	
	\section{Background \& Related Work}
	% Federated learning, drawbacks
	\subsection{Federated Learning}
	Federated learning (FL) is a training algorithm that allows a model to be trained from multiple sources in a decentralized manner \cite{https://doi.org/10.48550/arxiv.1902.01046}. In a case where the data is distributed among multiple locations such as crop data among farms, participants can obtain a global model that performs better than if they were trained solely locally \cite{DURRANT2022106648}. In FL, the locally trained parameters are sent to a trusted central server to perform FedAvg \cite{DBLP:journals/corr/McMahanMRA16}, which computes the average weights of local models. The updated parameters are then sent back to the local clients and this iterative process continues until convergence is reached. The updates do not contain the raw data but only the model information, which is not a direct representation of the data itself \cite{bonawitz2016practical}. As the data stays with the owner and never leaves the local devices, FL is often used as a measure to preserve sensitive information that data owners are unwilling to disclose. The ultimate objective of FL is to find a set of parameters, represented by $\theta_{global}$ that minimize the global loss function $\mathcal{L}$ across all clients, denoted as $k$, such that:
	\begin{equation} \label{eq1}
		\begin{split}
			\theta_{global} \in \argmin_{\theta} \mathcal{L}(\theta): = \frac{1}{k} \sum_{i}^k \ell_i (\theta)
		\end{split}
	\end{equation}

	During training, the server and clients need to communicate frequently to exchange updates. This communication incurs a cost in terms of data transmission. As contemporary models have shown tendency to exponentially grow in size \cite{qiu2020pre, menghani2023efficient}, a huge communication cost in federated learning can be expected, and it leads to heavy overheads on clients and high environmental burdens \cite{Wu_2022}. In a practical setting, the clients often communicate at different intervals due to their different hardware capabilities, size of the dataset (more data takes longer to train), and communication bandwidth. The work of those who struggle to communicate is often dismissed if they are unable to match the pace with their peers \cite{10.1145/3434770.3459739}. It may be less than ideal to aggregate without waiting for other clients, especially in the presence of heterogeneous data. Yet waiting for the slow clients can slow down training \cite{wang2020tackling}. 
	
	Some methods directly compress the size of updates by reducing the number of parameters.  FedKD \cite{Wu_2022} and FD\cite{jeong2018communication} compress updates with knowledge distillation, which trains smaller models to replicate the performance of large models. Konečný et al. transforms the structure of local updates into multiplications of smaller matrices and also compresses updates with lossy compression \cite{konevcny2016federated}. Zhu and Jin consider the minimization of communication cost and the maximization of global accuracy as two objectives and use a multi-objective algorithm to optimize both simultaneously \cite{zhu2019multiobjective}. The STC framework uses a series of techniques such as sparsification, ternarization, error accumulation and optimal Golomb encoding to compress the updates \cite{sattler2019robust}. LotteryFL sends sparse networks containing fewer parameters \cite{9708944}. There are also methods that indirectly reduce communication costs via optimizing the update schemes \cite{caldas2018expanding} and excluding the less influential devices in communication every round. eSGD updates based on the importance of the client - uploading local parameters only if the local loss is better than the previous round \cite{tao2018esgd}. DGC compresses the gradient while employing momentum correction, local gradient clipping, momentum factor masking and warm-up training to preserve the performance \cite{lin2020deep}.

	Another common obstacle in FL is having dissimilar data distribution or data heterogeneity among the silos, i.e., each silo contains non-identical and independent data (non-IID). In an IID scenario, the local gradient $\nabla$f is an approximation of the global $\nabla$F, which implies that the global model should also provide an unbiased inference for each local client. Much of related literature is based on the assumption of IID data distribution. However, in many cases, the training data present on the individual clients is collected by the clients themselves based on their local environment and usage pattern, and both the size and the distribution of the local datasets will typically vary heavily between different clients \cite{https://doi.org/10.48550/arxiv.1903.02891}. Models trained federatively in the present of such non-IID data may display significantly lower accuracy, and at times experience difficulties for the model to converge \cite{10.1145/3434770.3459739}. The reduction in performance can be explained by the weight divergence, where the non-IID data causes the global model to shift away from the ideal model (model obtained if data were IID), and this divergence may increase over communication rounds resulting in worse convergence and performance \cite{ZHU2021371}. Methods such as FedProx \cite{https://doi.org/10.48550/arxiv.1812.06127}, VRL-SGD \cite{https://doi.org/10.48550/arxiv.1912.12844} and SCAFFOLD \cite{https://doi.org/10.48550/arxiv.1910.06378} have been proposed to address the problem, but they either result in more communication cost or require longer to train.
	
	\subsection{Pruning}
	% pruning
	Techniques for eliminating redundant parameters have been widely used to make machine learning models smaller and more efficient \cite{https://doi.org/10.48550/arxiv.1510.00149, https://doi.org/10.48550/arxiv.1608.08710, NIPS1989_6c9882bb}. Pruning is one of the most established methods to make a neural network deployment and inference more efficient. Early literature commonly theorizes that pruning helps generalize a model \cite{lecun1989optimal}, and it reflects the classical understand of the bias-variance trade off. A model may result in sub-optimal performance if it is overparameterized (too complex) or underparameterized (too simple) \cite{rasmussen2000occam}. This can be observed in more recent findings - in fact, for small amounts of compression, pruning can increase accuracy \cite{https://doi.org/10.48550/arxiv.1506.02626, suzuki2018spectral}. Contemporary work on pruning has been largely motivated by the benefits that 1) it can reduce the energy required to run such large networks so they can run in real time on mobile devices, and 2) a reduction in model size from pruning also facilitates storage and transmission of mobile applications incorporating Deep Neural Networks (DNN) \cite{https://doi.org/10.48550/arxiv.1506.02626}. For instance, sparse neural networks can be made more power efficient and faster through a hardware accelerator such as EIE \cite{han2016eie}. 
	
	Deep networks are often so overparameterized that some neurons or connections do not improve the accuracy. This redundancy can be removed to transform into a sparse network with fewer parameters than the original without compromises on the performance. For instance, pruning can reduce the number of parameters in AlexNet by 9x and in VGG-16 by 13x on ImageNet, without any drop in predictive performance \cite{https://doi.org/10.48550/arxiv.1506.02626}. Pruning can be formulated as the following, where $x$ is the data and $\theta_p$ is the pruned weight, and we find a number of non-zero parameters $||\theta_p||_0$ to be smaller than the threshold $N$:
	\begin{equation} \label{eq prune}
		\begin{split}
			\argmin _{\theta_p} \mathcal{L}(x; \theta_p)\\
			\text{subject to: } ||\theta_p||_0 < N
		\end{split}
	\end{equation}

	There are many methods of producing a pruned model from a full network, and nearly all of them are derived from a well-established algorithm by Han et al \cite{https://doi.org/10.48550/arxiv.1506.02626}. It employs a three-step process, which begins by training a model to learn the importance of the weight connections. The second step is to prune the low-importance weights. Although there are many different criteria \cite{lecun1989optimal, hassibi1993optimal, li2016pruning} that can be used to determine what weights to prune, a simple yet effective method that is widely used today is the magnitude criterion where the importance of weights is determined by their absolute values ($importance = |\theta|$). The accuracy of the network is typically reduced immediately following pruning. The final step retrains (or fine-tunes) the remaining weights to recover. This process can be performed iteratively to gradually reduce the size of the network until the target sparsity is reached. It has been repeatedly shown that pruning randomly gives worse performance than following certain strategies \cite{yu2018nisp,gale2019state,https://doi.org/10.48550/arxiv.1803.03635}. Similarly, applying the same pruning ratio to all layers may result in worse performance than dynamically adjusting for each layer \cite{https://doi.org/10.48550/arxiv.1506.02626, gale2019state, https://doi.org/10.48550/arxiv.1608.08710}. A simple and effective way is to prune globally \cite{https://doi.org/10.48550/arxiv.1803.03635, lee2019snip} or alternatively find the optimal compression settings with an automated tool such as AutoML \cite{he2019amc}. Lottery Ticket Hypothesis (LTH) shows that it is possible to train a sub-network from scratch instead of fine-tuning the remaining weights. These sub-networks are discovered by resetting the weights to initialization after pruning \cite{https://doi.org/10.48550/arxiv.1803.03635}, and they are also called a ticket since they are hard to find.

	%%%%%%%%%%%%%%%%%%%%%%%%%%%%%%%%%%%%%%%%%%
	\section{Materials and Methods}
	
	\subsection{Data}
	Our demonstration focuses on collaborative federated forecasting using an accessible open-source dataset, given limited real data availability in agri-food. We use the tabular data anaylzed from a previous work of Khaki et al. \cite{Khaki_2020} for the same task of predicting the yield of soybean (bushels per acre). The dataset is composed of weather, soil and management data of soybean from 9 states and their counties from 1980 to 2018. In order to maintain consistency with the previous study, we use data from 1980 to 2015 to predict yield for the final three years 2016, 2017, and 2018.
	
	\begin{itemize}
		\item The weather data includes the weekly average of 6 attributes: precipitation, solar radiation, snow water equivalent, maximum temperature, minimum temperature and vapour pressure. This data was acquired from Daymet \cite{https://doi.org/10.3334/ornldaac/1840}.
		\item The soil data includes 10 attributes: bulk density, cation capacity exchange capacity at 7 pH, course fragments, clay percentage, total nitrogen, organic carbon density, organic matter percentage, pH in water, sand, silt, soil organic carbon, all measured at 6 depths. The data was acquired from Gridded Soil Survey Geographic Database \cite{gssurgo_2023} for the United States and is generally the most detailed level of soil geographic data in accordance with the national survey standards.
		\item The management data includes the cumulative percentage of planted fields within each state. This data is acquired from National Agricultural Statistics Service of the United States \cite{usda-nass_2019}.
	\end{itemize}
	
	The soil data is uniform throughout the period for each county while the weather and management data change over time. The data is distributed into 9 silos, with each representing a corresponding state. After cleaning and data processing (following mostly same procedure as \cite{Khaki_2020}), silos containing a wide range of samples are resulted as shown in Table \ref{tab data}. This imbalance in data distribution happens frequently in practical scenarios. Clients with a large training sample size account for a larger proportion of the overal training data, and it can reduce the accuracy of the minor clients \cite{5128907}. Data augmentation \cite{TANN1987} is a technique to increase the diversity of training data and reduce data imbalance issues. We applied random oversampling by making all silos equal in sample size. It does not cause any loss of information and alleviates the imbalance by replicating the minority samples more and majority samples less, resulting in equal distribution among the silos.

	\begin{table}[!htbp]
		\centering
		\caption{Data sample breakdown for the states used for prediction year 2016, 2017 and 2018. Each state represents a silo used in FL.\label{tab data}}
		\begin{tabular}{*7c}
			\toprule
			Location &  \multicolumn{2}{c}{2016} & \multicolumn{2}{c}{2017} & \multicolumn{2}{c}{2018}\\
			\midrule
			{}   & Train   & Val    & Train   & Val   &Train  &Val\\
			\hline
			Illinois   &2977    &67   &3044   &72   &3116   &64\\
			Indiana   &2630    &52   &2682   &54   &2736   &61\\
			Iowa   &3132    &94   &3226   &90   &3316   &86 \\
			Kansas  &2443    &17   &2460   &19   &2479   &15\\
			Minnesota   &2134    &55   &2189   &55   &2244   &43\\
			Missouri    &2395    &18   &2413   &19   &2432   &15\\
			Nebraska    &2274    &52   &2326   &43   &2369   &39\\
			North dakota    &574    &12   &586   &12   &598   &12\\
			South dakota    &1164    &22   &1186   &21   &1207   &20\\
			\hline
			Combined    &19723  &389    &20112  &385    &20479  &355\\
			\bottomrule
		\end{tabular}
	\end{table}

	\subsection{Federated Pruning}
\begin{figure}[H]
		\begin{adjustwidth}{-\extralength}{0cm}
			\hfill
			\includegraphics[width=17.8 cm]{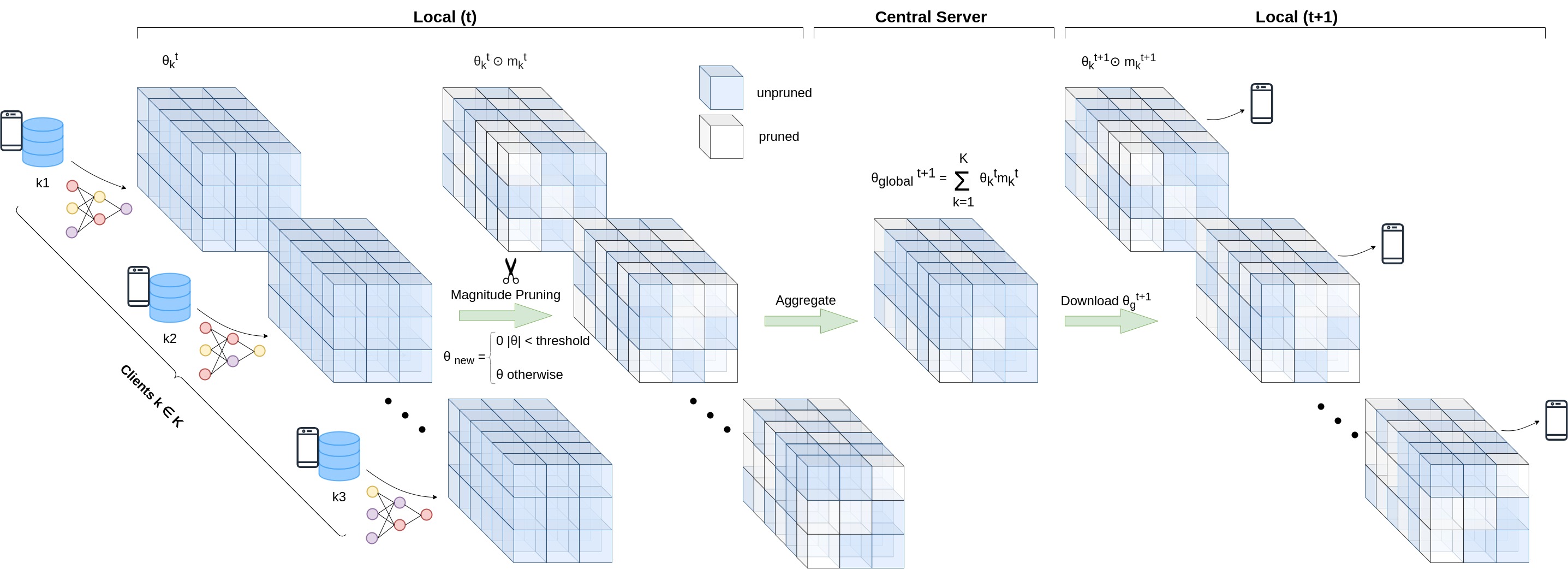}
		\end{adjustwidth}
		\caption{Federated Pruning during one round of federated learning.\label{fig overview}}
	\end{figure}

	Despite the theoretical benefits, FL may face challenges in learning an optimal global model for all clients. In this paper, we propose a method that improves upon FL through the use of pruning. Figure \ref{fig overview} presents an overview of our method. Instead of training for a model that is globally shared among all silos, we focus on the individual client models. Pruning local models results in models with unique sets of parameters. They become localized in the process as the remaining weights are most important specifically for their own data distribution. Pruned models are also smaller in size and require less data for communicating with the server. Reducing data transmission means the models can be uploaded to the server more quickly, and result in lower latency for training. This could be especially helpful in agri-food settings such as farms and rural areas where the internet connectivity and bandwidth are more limited. Once the sparse models are uploaded, the global server aggregates them and updates clients with the new weights. Like FL, this process is iterated for T communication rounds. Localized and compact models are produced by the end of the process. We name this method Federated Pruning (FedPruning) throughout this paper. FedPruning can be described as in Equation \ref{eq fedpruning}. $||\theta_p||_0$ the L0 norm of the pruned parameters $\theta_p$, representing the number of non-zero elements. To obtain the pruned network $\theta_p$, we prune with the pruning function $P$ if $||\theta_p||_0$ is less than a preset threshold $N$. We obtain client parameters $\theta_k$ by averaging client $\theta_p$ in aggregation during federated learning.
	
	%from heterogeneous data across the silos and communicate efficiently between the central server and the clients due to network constraints in practical agri-food settings. In our work, we propose an algorithm that could mitigate these challenges through the use of pruning. Given a number of clients and their local datasets, our proposed approach learns localized, communication-efficient and compact models, which are suitable for the deployment of edge devices with limited computational resources and storage capacity. 

	\begin{equation} \label{eq fedpruning}
		\begin{split}
			\theta_{k} \in \argmin_{\theta_p} \mathcal{L}(\theta_p): = \frac{1}{k} \sum_{i}^k \ell_i (x;\theta_p)\\
			\text{where } \theta_p = P(x; \theta) \text{ s.t. } ||\theta_p||_0 < N
		\end{split}
	\end{equation}
	
	\subsection{Localization-preserving Aggregation}
	
	The primary incentive to participate in FL is to have a global model that is better performing than the individual local models. However, in practice, local clients can outperform the global model due to data heterogeneity \cite{DBLP:journals/corr/abs-2002-05516} or not independent and identically distributed (Non-IID) data, and it defeats the purpose of FL. For a supervised learning task on client $k$, the data is in the form of (X,y) where x is the input features and y is the label, and it follows a local distribution $P_k(X,y)$. By Non-IID, the $P_k(X,y)$ differs from client to client \cite{ZHU2021371}. We may experience this from different types of skews. First, the conditional distribution $P_k(y|X)$ may be different across the clients although $P_k(X)$ is the same. In the agri-food sector, this could be resulted from the different measuring devices or sensitivities of sensors used to capture the local data. Second, for time-series data such as ours, it can happen when clients have uneven distribution across the years. Some may have more data points towards later years while others have more from the early years. 
	
	While the widely popular FedAvg can work with non-iid to some degree, it ultimately lacks the theoretical guarantee to converge for all clients \cite{Li_2020}. During communication rounds, it aggregates the gradients of the local models by taking the weighted average of the local gradients and returning it back to the clients \cite{DBLP:journals/corr/McMahanMRA16}. It results in handling all the various data distributions with one single global model. In our method, instead of attempting to obtain a "one model to fit all", each client learns a localized sparse network that is tailored to its own data distribution.
	
\begin{figure}
		\centering
		\includegraphics[width=9cm]{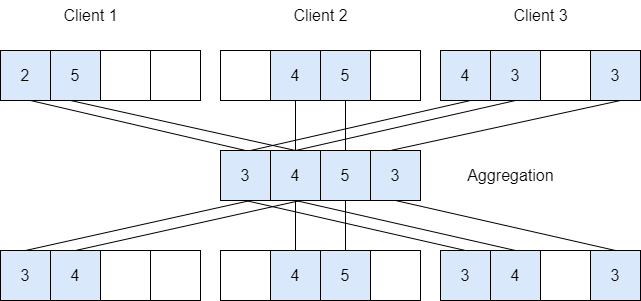}
		\caption{Localization-preserving aggregation.\label{fig1}}
	\end{figure}

	Following the idea of FedAvg where the average weights are proportional to the number of participants, we use the aggregation algorithm proposed in LotteryFL \cite{9708944}, which is designed to aggregate sparse networks. Largely pruned networks may overlap with each other on some connections while having unique or infrequently overlapped connections due to the non-IID data distribution across the clients. This aggregation strategy updates on only the overlapped connections of the sub-networks while keeping the non-overlapped parts unchanged. Figure \ref{fig1} illustrates how the averages of sparse tensors are computed. For instance, at the leftmost position, we take the average from client 1 and client 3 since client 2 has it pruned. The aggregated weights are then sent back to the clients and the unpruned weights are updated. This aggregation strategy allows the server to maximally preserve weights that are important to individual clients.
	
	\subsection{Communication Cost}
	In the field of agri-food, a persistent issue is the insufficient availability of robust wireless connectivity in rural regions. Any delay in data exchange or loss of connection among IoT devices such as sensors or electronic devices could directly hinder farming operations \cite{VANHILTEN2022107291}. The application of sparsity to neural networks is a widely adopted technique for minimizing the number of parameters transmitted to the server, and thereby reducing communication costs \cite{Aji_2017, https://doi.org/10.48550/arxiv.1809.10505}. Communication constitutes a common bottleneck in FL since the participating clients are regularly required to send and receive updates from the server. By removing a portion of parameters per update, the communication cost is reduced to the compact size of a sub-network from the full network.
	
	% In the domain of technology applications within the agri-food sector, particularly in the context of IoT, wireless communication necessitates low power consumption. In our work, we evaluate the real-time performance of deploying localized models on the Nvidia Jetson platform, including including memory, latency, and power consumption.
	
	\subsection{Iterative Magnitude Pruning}
	We obtain a sparse neural network on each client $k\in K$ by training the network and pruning its smallest-magnitude weights. Consider a dense neural network $f(x;\theta_k)$ with parameters $\theta_k$, when optimizing with gradient descent on a training set, the validation loss $l_k$ converges. In the implementation, pruning a percentage of the weights leads to the generation of a mask $m_k \in \{0,1\}$. If the magnitude of parameter is smaller than the quantile, the corresponding mask entry is set to 0. The mask is combined with the state of the network to produce a sub-network $f(x;\theta_k \odot m_k)$, which is then trained again to recover $l_k$ (If LTH is used, the weights of the sub-network is reset to $\theta_g$ at initialization before re-training). The network is trained and pruned over $T$ rounds; each round prunes $p\%$ of the remaining weights that survived the round. This iterative magnitude pruning step makes the basis of the client updates presented in our approach.
	
	We define the sparsity of a network as $P_m = \frac{||m||_0}{|\theta|}$, with $||m||_0$ being the number of zeros in a mask and $|\theta|$ being the number of weights in a network, i.e., $P_m$ = 75\% means that 75\% of weights have been pruned. $P_m$ is used as the metric to evaluate the compactness of the models throughout this paper. Our approach produces sub-networks $\exists m$ for which $l' \approx l \textit{ (comparable losses)}$ and $||m||_0 \ll \theta \textit{ (fewer parameters)}$ for all clients $K$.

	\subsection{Training Algorithm}
	The general training algorithm is formally described in Algorithm \ref{Alg1}. The steps are taken as follow:
	
	\begin{enumerate}
		\item Global weights are downloaded from central server to clients.
		\item Perform local training on each client for $e$ epochs.
		\item If $\textit{loss} < loss_{best}$, prune $p\%$ of the parameters $\theta$ by magnitude, generating mask $m$. Train for $e$ epochs to recover the loss.
		\item The current sub-networks $\theta_k^t$ of the clients are sent to the global server for aggregation. Update the global model $\theta_k^t$ to $\theta_k^{t+1}$.
	\end{enumerate}
	The above steps are iterated for $T$ rounds. We use magnitude as our pruning criterion to determine the importance of weights. To give networks enough time to converge, we avoid pruning prematurely by running regular FL without pruning for 2 communication rounds - we find that it greatly reduce the risks of networks failing to recover, and increases the pruning potential in future rounds. 
	
	In our test setting, clients converge at different rates and their eventual losses vary. Thus, we perform pruning based on their own convergence and use their best loss as an indicator for what they are ready to be pruned. We ensure that the loss recovers to its previous best before it can be pruned again. In practice, we set this threshold to 10-20\% over its best loss to accelerate pruning early on as it is likely that the loss will recover in later rounds. Obtaining smaller models early on means that there is less communication cost throughout the entire FL training process. To allow losses to recover and stabilize following pruning, a number of non-pruning round is added. 
	
	\begin{algorithm}
		\caption{Training Algorithm. T rounds are indexed by t; The clients are indexed by $k$; r is the prune ratio; m is the local mask; $\eta$ is the learning rate;  $\ell$ is the loss function }
		\label{Alg1}
		\begin{algorithmic}[1]
			\State initialize global model $\theta_{global}$ with $\theta_0$ 
			\While {$\textit{round t < T}$}
			\State \textbf{ClientUpdate($\theta_{global}):$}
			\For {$\textit{each client k1, k2}\ldots$}
			\State {$\theta_k^t \gets \theta_{global} \odot m_k^t$}
			\State {$\theta_k^t$ $\gets$ Train($\theta_k^t$)}\\
			\algorithmiccomment{If loss has recovered; target sparsity hasn't reached; no pruning in final rounds}
			\If {$loss < loss_{best} \textit{ and } r^t < r_{target} \textit{ and } t<T-3 $}     
			\State {$m_k^t \gets \textit{Prune p\% of }\theta_k^t$}
			\State $\theta_k^t \gets \theta_0$ \algorithmiccomment{Reset weights to initialization (if LTH is used)}
			\State {$\theta_k^t$ $\gets$ Train($\theta_k^t$ $\odot$ $m_k^t$)}   \algorithmiccomment{recover loss}
			\EndIf
			\State Return $\theta_k^t$ to server
			\EndFor
			\State \textbf{Server Executes:}
			\State {$\theta_{global}^{t+1}$ $\gets$ aggregate($\theta_{k1}^{t}, \theta_{k2}^{t}, \dots$)}   \algorithmiccomment{Note: client weights have already been masked}
			\EndWhile
			
			\Function{Train}{$\theta_k^t$}
			\For {$\textit {epoch e}=1,2,\dots, $}
			\For{$\textit{batch b} \in \textit{B}$}
			\State $\theta_k^{t+1} \gets \theta_k^t - \eta \nabla\theta_k^t\ell(\hat{y}, y)$ 
			\State \Return $\theta_k^{t+1}$
			\EndFor
			\EndFor
			\EndFunction
		\end{algorithmic} 
	\end{algorithm}

	\subsection{Model Architecture}
	The previous work by Khaki et al. \cite{Khaki_2020} tested five different models for this forecasting task - CNN-LSTM, Random Forest, Deep fully connected neural network (DFNN) and LASSO. The CNN-LSTM was most effective in predicting yields of soybean with RMSE for the validation data being approximately 8\%. However, the line of research for iterative magnitude pruning is almost exclusively based on convolutional and fully connected layers \cite{https://doi.org/10.48550/arxiv.2101.09671}, so we build a model based on these layers to realize the theoretical benefits of pruning. We utilize the previously established convolutional layers to capture the temporal structure of the data for weather, soil and management, but concatenate them along with yield data from the previous dependent years into the fully connected layers, which work as our regressor. Batch normalization is also used after the non-linearity to accelerate training and improve the accuracy \cite{https://doi.org/10.48550/arxiv.1805.11604}. We experimented with different configurations and found that 3 fully connected layers gave an accuracy comparable to the DFNN model from the previous work. Rectified linear unit (ReLU) activation function is used for all convolutional and fully connected layers. %The model configuration is summarized in Appendix Table \ref{tabb}.

	%%%%%%%%%%%%%%%%%%%%%%%%%%%%%%%%%%%%%%%%%%
	% This section may be divided by subheadings. It should provide a concise and precise description of the experimental results, their interpretation as well as the experimental conclusions that can be drawn.
	% \subsection{Subsection}
	% \subsubsection{Subsubsection}
	
	\section{Results}
	\subsection{Experiment Setup}
	To evaluate the inference accuracy of the models, the data for 2016, 2017 and 2018 are used as validation years and their yields are predicted in bushels per acre. We implemented three baselines to make a fair comparison.
	\begin{itemize}[labelsep=2.5mm,topsep=-3pt]
		\item \textbf{Centralized} is trained on the combined data of the clients.
		\item \textbf{Local only} is trained locally by each client.
		\item \textbf{FedAvg} is the classic FL approach where clients download the global model from the server, train using local data, and then send updates to the server to update the global model through aggregation.
	\end{itemize}\vspace{3pt}
	
	To make local models more accurate and compact, each sub-network prunes the least important parameters after local training. The hypothesis behind this is that the surviving parameters are those that are most important to the local clients (not other clients), and the aggregation of pruned models will preserve the localized parameters derived from pruning. To evaluate the effectiveness of our proposed method, we implement four variations, and compare them with the baselines.
	\begin{itemize}[labelsep=2.5mm,topsep=-3pt]
		\item \textbf{Federated Pruning (FedPruning)} is the method described in Algorithm \ref{Alg1} where individual models are pruned on the client's end and sub-networks are aggregated on only the overlapped connections on the server's end. The surviving sub-networks are fine-tuned and their losses are recovered. 
		\item \textbf{Federated Pruning with Lottery Ticket (FedPruning-LT)} is trained on the same algorithm as FedPruning, but with LTH \cite{https://doi.org/10.48550/arxiv.1803.03635}, which resets the remaining weights to initialization, creating a winning ticket. Both FedPruning approaches prune iteratively, which repeatedly train, prune, re-train, aggregate over T rounds. Each round prunes $p\%$ of the survived weights until target sparsity is reached.
		\item \textbf{One-shot} prunes all client networks by a large $p\%$ at once \cite{lee2019snip} and proceeds with federated learning with the same aggregation strategy as FedPruning. 
		\item \textbf{One-shot with Lottery Ticket (one-shot-LT)} executes a one-shot pruning, but with weights reset to initialization following pruning.
	\end{itemize}\vspace{3pt}
	
	For all methods, weights are initialized with the Kaiming method \cite{https://doi.org/10.48550/arxiv.1502.01852}. Adam optimizer is used with a learning rate decaying at round 5 and 10 by a factor of 0.2. L2 regularization can be used to enforce sparsity during training by encouraging smaller weights \cite{https://doi.org/10.48550/arxiv.1506.02626}, and therefore we set a weight decay of 0.0001. For federated methods, the model is trained for a maximum of 40 communication rounds with 5-8 epochs trained locally. Local early-stopping is also used to prevent overfitting. Table \ref{tab1} includes additional parameter settings used in this experiment.
	
	\begin{table}[H] 
		\caption{Settings tested in this experiment. The pruning rate denotes the percentage of weights pruned each time pruning is executed, and target sparsity represents the intended level of sparsity achieved at the end of training. Notably, the sparsity achieved by iterative pruning may slightly deviate from the target sparsity depending on the training performance.\label{tab1}}
		\newcolumntype{C}{>{\centering\arraybackslash}X}
		\footnotesize
		\begin{tabularx}{\textwidth}{CCCCCCCC}
			\toprule
			\textbf{Method}    &\textbf{Centralized}    & \textbf{Local only}    &\textbf{FedAvg}    &\textbf{FedPruning}   &\textbf{FedPruning- LT}   &\textbf{One-Shot}   &\textbf{One-Shot-LT}     \\
			\midrule
			Rounds/ Local Epochs &1/300  &1/300  &40/5  &40/6   &40/8   &40/5   &40/5\\
			\hline
			Learning Rates   &5e-5, 1e-5, 0.2e-6   &2e-5, 4e-6, 8e-7   &2e-5, 4e-6, 8e-7    &2e-5, 4e-6, 8e-7    &2e-5, 2e-6, 4e-7   &2e-5, 4e-6, 8e-7    &1e-5, 2e-6, 4e-7   \\
			\hline
			Pruning Rate  &0.0    &0.0  &0.0  &0.25   &0.415   &0.70    &0.70 \\
			\hline
			Target Sparsity &0.0  &0.0  &0.0  &0.80    &0.80   &0.70    &0.70 \\
			\bottomrule
		\end{tabularx}
	\end{table}
	
\begin{figure}[H]
		\centering
		\includegraphics[width=14 cm]{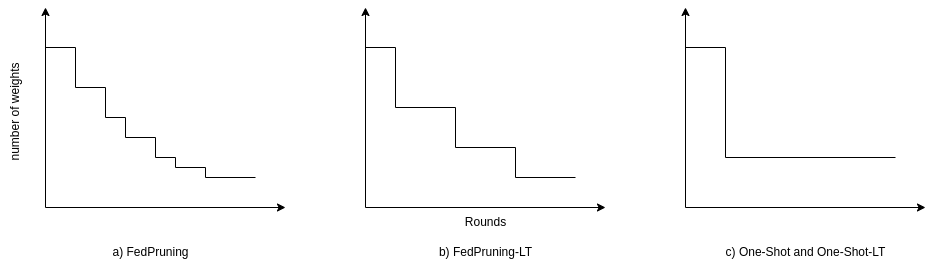}
		\caption{Sparcification Schedules.\label{fig schedule}}
	\end{figure}

	\noindent\textbf{Sparsification Schedules}. Model sparsity is trained with a schedule. As shown in Fig \ref{fig schedule}, FedPruning (a) prunes a relatively small percentage iteratively, whereas FedPruning-LT (b) prunes a maximum of three times but a larger portion each time. This is because in our testing, LT requires more iterations to recover due to the weight reset. If a network is pruned prematurely before the loss is recovered, it can impede its ability to ever recover its former loss and also prevent further pruning. Hence, FedPruning-LT is given a minimum of 7 rounds to recover following a pruning round, while FedPruning is given a minimum of 3 rounds to recover. In addition to these mandatory recovery rounds, clients are evaluated before pruning and pruning is only executed if their losses have been recovered. Pruning is also prohibited in the final rounds to ensure the best final inference performance. Both one-shot approaches (c) prune a significant portion as soon as the network converges, and proceed with training and aggregating sparse networks.

	\subsection{Inference Performance, Communication Cost and Size}
	The centralized baseline pools all training and testing data together, and demonstrates a commensurate level of inference performance when compared to the DFNN model from the previous study \cite{Khaki_2020}. We find that the local models, which only use data from their respective silos, exhibit poor performance compared to the centralized baseline, as indicated in Table \ref{tab rmse}. This is expected as the local clients have limited training data. FedAvg addresses the data limitation by leveraging all data from all silos via aggregation and model updates. We evaluate the performance of FedAvg using the global model directly following the aggregation. Our findings indicate that FedAvg outperforms the local models for all years by approximately 10.5\%. Nevertheless, FedAvg still falls short of the centralized models in terms of performance.
	
	Since FedPruning produces different, localized models for each client, we evaluate the individual client models instead of the global model like we do with FedAvg. Therefore, to evaluate the performance of FedPruning - both with iterative and one-shot pruning, with and without LTH, we assess the inference performance using local test data from each silo at the end of local training. The most direct way for the method to be demonstrably useful is for it to be a drop-in replacement for FedAvg - that is it must be able to reach the inference performance no worse than FedAvg on all clients on average, and result in fewer parameters within a specific number of communication rounds. Symbolically, $\mathcal{L}(\mathscr{A}_\text{fp}^{0\rightarrow T}(\theta_p)) \leq \mathcal{L}(\mathscr{A}_\text{fa}^{0\rightarrow T}(\theta))$  and $|\theta_p| < |\theta|$, where $|\theta_p|$ and $|\theta|$ are the numbers of parameters in pruned and unpruned models respectively, $\mathcal{L}$ is the average loss across all clients, $\mathscr{A}_\text{fp}^{x\rightarrow y}$ and $\mathscr{A}_\text{fp}^{x\rightarrow y}$ are the FedPruning and FedAvg procedure for training from round x to round y, and T is the final round.

	\noindent\textbf{Assessing the inference performance of local models.} The results indicate that FedPruning under all four settings significantly outperform FedAvg, with performance improvements ranging between 15.5\% to 19.8\%. The iterative pruning methods FedPruning and FedPruning-LT outperform the centralized baseline for 2017 and 2018, while exhibiting slightly inferior performance for 2016. Overall, these methods demonstrate comparable inference performance to the centralized baseline. One-shot surprisingly shows a similar performance compared to its iterative counterpart, despite being slightly less pruned. However, when LTH is applied to one-shot, it shows a small but noticeable increase in performance across all years, amounting to a 5.5\% improvement overall. We also make the observation that these variations collectively are on par with the centralized baseline or even marginally outperform it. One-shot-LT shows the biggest difference by approximately 10\%.
	
	\begin{table}[H]
		\caption{RMSE(bushels per acre) of the 9 states trained using different training procedures. The values are recorded using the average of 3 runs each year with random initialization seeds. The final average sparsity for FedPruning and FedPruningLT are included in the bracket. The sparsity for one-shot and one-shot-LT are 0.7. \label{tab rmse}}
		%\begin{adjustwidth}{\extralength}{-2cm}
		\newcolumntype{C}{>{\centering\arraybackslash}X}
		\resizebox{\columnwidth}{!}{\begin{tabularx}{\fulllength}{CCCCCCCC}
				\toprule
				\textbf{Year}	& \textbf{Centralized Baseline}	 & \textbf{Local Only}  & \textbf{FedAvg} & \textbf{FedPruning}    &\textbf{FedPruning -LT}    &\textbf{One-shot}  &\textbf{One-shot -LT}\\
				\midrule
				2016		    &8.74	& 10.10  &9.70  &8.87 (0.75) &9.52 (0.73) &8.50 &8.39\\
				2017            &6.04   & 7.46   &6.53   &5.02 (0.84) &5.24 (0.76) &5.54 &4.85\\
				2018            &5.83   & 7.96   &6.85   &5.62 (0.78) &5.52 (0.76) &5.55 &5.26\\
				\\
				Average         &6.87   &8.84   &7.69   &6.50 (0.79)  &6.75 (0.75)  &6.53  &6.17\\
				\bottomrule
		\end{tabularx}}
		%\end{adjustwidth}
	\end{table}
	
\begin{figure}[H]
		\centering
		\begin{subfigure}[b]{0.52\textwidth}
			\includegraphics[width=\textwidth]{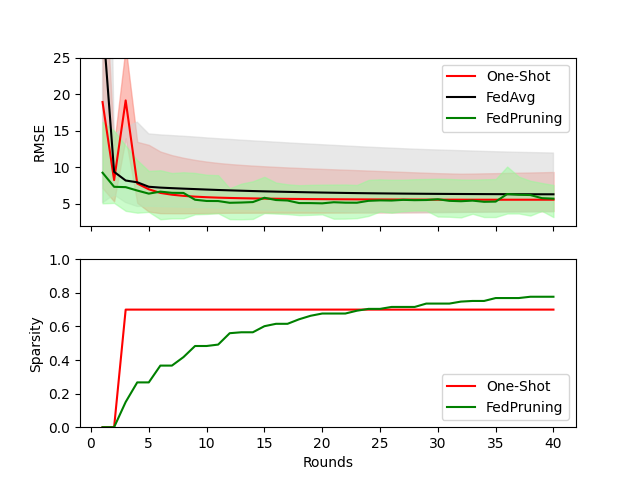}
			\label{fig:f1}
		\end{subfigure}
		\hspace{-1cm} % <-- Adjust the horizontal spacing between the subfigures
		\begin{subfigure}[b]{0.52\textwidth}
			\includegraphics[width=\textwidth]{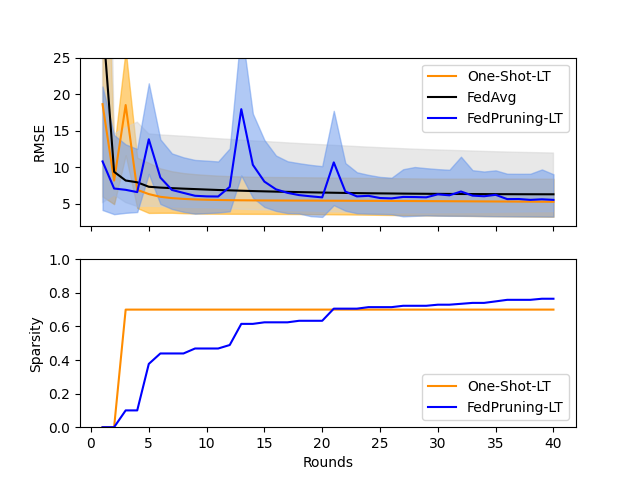}
			\label{fig:f2}
		\end{subfigure}
		\caption{RMSE performance and model sparsity of FedPruning, FedPruning-LT, One-Shot, and One-Shot-LT compared to the FedAvg basedline over 40 communication rounds for 2018. The solid lines at top represent the average RMSE of clients. The shades correspond to the RMSE between the highest and lowest clients. The graph is produced using the average of 3 runs with random initialization seeds. \label{fig history}}
	\end{figure}

	Figure \ref{fig history} shows how the pruning approaches perform in terms of RMSE and sparsity over 40 communication rounds. It is observed that FedPruning and One-shot (left graph) have a narrower client RMSE range than FedAvg, with the upper bound of client RMSE being lower. Additionally, FedPruning exceeds the final performance at a lower sparsity (approximately $P_m = 30\%$ - 60\%) and decreases slightly as we prune, forming Occam's Hill, which suggests that if the model is either too simple or too complex, performance on an independent test set will suffer \cite{rasmussen2000occam}. FedPruning-LT (right graph) prunes more each time but fewer times in total. We tested it with the same schedule as FedPruning, and observed that when client models are pruned before they recover, they may permanently lose the performance. This not only affects itself, but also other clients since some of their weights are shared. We also find that the overall performance may be better when the clients are pruned and recover together, as opposed to independently. The RMSE spikes on the graph indicate the impact of weight reset after pruning, and we allow them to recover fully before the next pruning round. Despite the effort, it does not appear to match the performance of FedPruning and One-Shot-LT. One-Shot-LT on the other hand, although simpler to implement, consistently shows superior performance compared to both one-shot and FedPruning-LT. 
	
	\noindent\textbf{Assessing the cost of communication and size of model.} The measurement of communication costs is commonly based on the magnitude of data volume transmitted between clients and server. It is influenced by two distinct factors. Firstly, the sparsity of a model, which affects its compression ratio during communication, plays a significant role in determining the quantity of data transmitted. Sparse models, being inherently smaller in size, require less communication cost to transmit as compared to larger models. Secondly, the timing of model pruning also significantly affects the communication cost. The earlier a model is pruned, the greater the amount of communication cost that can be saved in the long run. This can be visualized in Figure \ref{fig history}. Specifically, it depicts the data consumption patterns of the iterative and one-shot approaches. As per the figure, client models that prune iteratively consume a relatively greater amount of communication cost initially and gradually reduce it as they become smaller. In contrast, one-shot approaches prune in one go and achieve a steady reduction in communication cost. All pruning methods have demonstrated the ability to effectively reduce model size while preserving inference performance. Notably, one-shot pruning methods have been observed to achieve a size reduction of 3.22X, while FedPruning and FedPruning-LT have reduced model size by 4.76X and 4X, respectively.

	\begin{table}[H] 
		\caption{Communication cost and size of an average client model during the FL process. The values are recorded using the average of 3 years, with 3 runs per year with random initialization seeds. \label{tab2}}
		\newcolumntype{C}{>{\centering\arraybackslash}X}
		\begin{tabularx}{\textwidth}{CCCC}
			\toprule
			\textbf{Method} &\textbf{Communication Cost (MB)} &\textbf{Communication Saved (\%)}	& \textbf{Client Model Size (KB)}	 \\
			\midrule
			FedAvg  		&50.76  &0      &634.64	\\
			FedPruning      &21.78  &57.1   &133.27 \\
			FedPruning-LT   &20.96  &58.7   &158.66 \\
			One-Shot        &17.92  &64.7   &196.95 \\
			One-Shot-LT     &17.92  &64.7   &196.95 \\
			
			\bottomrule
		\end{tabularx}
	\end{table}
	\unskip

	\section{Discussion}
	Existing works \cite{https://doi.org/10.48550/arxiv.1506.02626, https://doi.org/10.48550/arxiv.1803.03635} demonstrate that neural networks can be represented by substantially fewer parameters. We drew inspiration from the benefits of network pruning and implemented the fundamental pruning steps, which are training the initial network, removing connections, and fine-tuning the model for the participating local clients in the FL process. Subsequently, we aggregated the overlapping connections of the sub-networks instead of the full network. Through testing with the soybean yield prediction dataset, we observed that the proposed method consistently outperformed the benchmark of the classic FedAvg approach across all years tested. Additionally, this method efficiently produced more compact models for edge device utilization, thereby introducing new options for the implementation of FL in practical agri-food settings. These findings suggest that the proposed method could potentially improve the efficiency and feasibility of FL in agri-food settings, and may offer practical advantages over the classic FedAvg approach.
	\\
	
	\noindent\textbf{Model localization.} In theory, if all client datasets are identically and independently distributed, and the overall data volume is large enough, then FedAvg and the centralized model using combined data could achieve similar inference performance because the client stochastic gradient is an unbiased estimate of the full gradient and the average model weights of the client models will approximate the centralized model \cite{bottou2010large, rakhlin2012making}. However, this assumption nearly never holds in practice, as we can see from our experiment. The non-IID data happens in the presence of inconsistent data distributions when there is an attribute imbalance of the training data across clients due to perturbations. When the number of data collection points become large, it is difficult to keep the measurements consistent. For instance, the measurement of temperature may vary between farms due to the deployment of sensors in different positions within the polytunnels. Different farms may also use different fertilizers and be exposed to different climates, or elements not captured in the datasets but affecting the yield. By producing sparse neural networks, the weights important to the client itself are retained. As sparsity $P_m$ increases, the number of parameters shared with other clients via aggregation decreases. A percentage of these remaining weights are nearly unique or shared with few clients who also consider these weights important to themselves, and these parameters attribute to the localization of client models. A higher sparsity may reduce the divergence from non-IID data, which causes poor performance. This can be seen from our results, as FedPruning consistently outperform FedAvg on individual client models.
	\\
	
	\noindent\textbf{Pruning.} 
	For best results, rather than pruning all weights at once, the common practice is to repeat the train-prune-retrain procedure until the target sparsity is reached. As LeCun et al. \cite{lecun1989optimal} put "A simple strategy consists in deleting parameters with small "saliency",  i.e. those whose deletion will have the least effect on the training error... After deletion, the network should be retrained. Of course this procedure can be iterated. ". The concept of iterative magnitude pruning is also realized by many contemporary research works. Han et al. \cite{https://doi.org/10.48550/arxiv.1506.02626}, who modernized this method states "Our pruning method ... learns the network connectivity via normal network training... The second step is to prune the low-weight connections... The final step retrains the network... This step is critical. If the network is used without retraining, accuracy is significantly impacted.". Modern literature seems to agree that pruning should be performed iteratively for best inference performance. However, in our study, we did not observe a significant disparity between the iterative and one-shot approaches. This may suggest that the true potential of iterative FedPruning has yet to be realized. Local losses degrade after the models are pruned, and it is observed that they may not recover to their pre-pruning state before they are passed to the central server. Although they typically recover in subsequent rounds, passing these models for aggregation may result in 'hiccups', a temporary degradation in the overall loss. This can be observed from the jagged loss curve in Figure \ref{fig history} compared to the smooth curves of FedAvg and the one-shot methods. We have established a generalizable algorithm that effectively leverages pruning during FL. However, there may be unexplored approaches to identifying more optimal settings for iterative pruning, such as determining the ideal granularity, schedule, and other related factors."
	
	When we applied LTH to FedPruning, we observed that the weight reset (required to find the ticket) resulted in a significant deterioration (increase in the local loss), returning it to the initial training state. Unlike the isolated experiments by Frankle et al. \cite{https://doi.org/10.48550/arxiv.1803.03635}, our local models required weight sharing with each other. When local models performed weight resetting, retraining and averaging independently and frequently, it would adversely affect the aggregated weights of other clients, causing the overall loss to cease improving. We reduced the noise caused by this by changing the schedule of sparsification - nearly simultaneous pruning across all clients and less frequently. Our experiments demonstrated that applying LTH to the iterative FedPruning was arduous and less advantageous compared to the other variations, though still outperforming the FedAvg baseline. However, LTH was more effective when applied only once, as shown in the comparison between one-shot and one-shot-LT. This finding may suggest that the property of LTH reemerges when scheduling of sparsification becomes less of an issue. It invites a number of follow-up questions and may be explored empirically in future research.
	\\
	
	\noindent\textbf{Sustainability and Inference Efficiency.} Large models should be compressed to effectively participate in FL and fit on edge devices, as they require more computations, energy consumption and carbon footprint. The efficiency of machine learning inference is dictated by memory locality - if a large model cannot be held in on-chip storage (SRAM), references need to be made to access off-chip memory (DRAM), and accessing DRAM memory (640pJ) is significantly more energy-intensive than accessing SRAM memory (5pJ) \cite{horowitz}. When compared to the energy cost of 32-bit float multiplications (3.7pJ), memory locality dominates. Sparse models with compatible hardware or framework require less computations and data movement during inference than their dense counterparts, and forms a step towards creating sustainable solutions in agri-food and beyond. 
	% After all, fewer parameters means fewer operations that must be performed on the forward and backward passes. However, because the pruned parameters are stored as zeroes in sparse matrices (in our implementation as $\theta \odot m_k$ ), the number of surviving parameters is not proportional to the number of operations performed when the models are run without further support for sparsity. Hardware manufacturers and researchers have recognized the potential for sparsity to improve sustainability and inference efficiency, and they have developed tools and methods with varying degrees of support for sparsity. For instance, Nvidia tensorRT has support for fine-grained sparsity when every four-weight block has two or less pruned weights. Cerebras and Graphcore also claim that their chips can accelerate unstructured sparsity. CPUs are known to be more compatible with unstructured pruning, and NeuralMagic in particular is exploiting this. Han et al. proposed EIE \cite{han2016eie} which compresses sparse neural networks making it 3,400x more energy efficient than a GPU compared at the time. Guo et al. proposes a sparsity pattern that can accelerate sparse models without any additional hardware support.
	\\
	
	\noindent\textbf{Use Cases.}
	As our goal is to propose technological solutions to facilitate data sharing and enable the development of efficient and 'green' machine learning models at scale, as well as to potentially encourage those in agri-food sector to adopt such technologies, we provide example use cases that could benefit from such a methodology, beyond the soybean case we are considering in this paper. Especially in areas where we see data sharing in agri-food via distributed training to be most applicable. In this paper, we focused on forecasting for collaborative federations for our empirical demonstration given the accessibility to suitable open-source datasets. However, the proposed methodology is directly applicable to the other use cases. We describe two key use cases observed in the agri-food sector that we believe data sharing and distributed/collaborative training with efficient machine learning models can assist; this list is not exhaustive.
	\begin{itemize}
		\item \textit{Strawberry yield forecasting for collaborative consortia} \\
		The aggregation of more data from a variety of sources and multiple farms can vastly improve the performance of machine learning models and other decision-support systems. In the agri-food sector, soft fruit growers, for instance, can be limited by their data collection processes, yet they may wish to employ decision-support frameworks to improve not only profits but also their sustainability (net-zero targets). Another aspect of this relates to contractual agreements between growers and large retail supermarkets; over-/under- estimating the amount of produce can lead to fruit waste or fruit shortages respectively, which can have both financial and environmental repercussions for growers and the sector. Having a technical solution in place that allows the creation of federations like that explored in \cite{DURRANT2021100493} facilitated through our proposed efficient federated learning methodology can enable multiple growers to share data in a trustworthy and transparent manner to improve their own processes and production systems towards achieving financial and environmental sustainability. A decentralized pruned model can also be deployed on edge and/or other devices for more efficient real-time inference.
		\item \textit{Plant diseases and pest detection from crop images}\\
		One of the most devastating factors that affect yield and the quality of plants relates to plant diseases and pests. Recognising early signs of such events is paramount towards damage limitation. Object recognition systems and remote monitoring can be useful tools that can help to identify such adverse events as early as possible, but they require lots of representative images to be used for training large-scale models. In practice, it might be unlikely that a single grower or farmer will have adequate data that can be used to train a single local model, performing well enough to be practically useful. However, aggregating multiple image datasets from various stakeholders that have been captured via single or multiple sources, e.g. infrared cameras, depth cameras, simple colour cameras, etc. can be transformational towards developing robust plant disease and pest identification systems that can recognise such problems early and help to reduce waste and contribute the financial viability of the growers and farmers, and the agri-food sector in general. Our proposed methodology can be used in such settings and be trained with multimodal data, therefore enabling decentralized training with efficient pruned neural network models. Such a lightweight model can be deployed on edge devices for real-time decision support.
	\end{itemize}
	
	\section{Conclusion}
	As machine learning models grow rapidly in size, they demand more memory and energy footprint, and make it especially challenging for FL on edge-devices in agricultural settings where network and hardware capacities are even more limited. Moreover, the non-IID data adversely affects the local accuracy of clients. We proposed a solution for communication-efficient federated learning that pruned the models locally and aggregated them at a global level. The advantages were the superior local inference as a result of the localized models, reduced model sizes for more efficient deployment and reduced communication costs during training. 
	
	To demonstrate the effectiveness of our method, we tested it with various pruning policies on a real agri-food dataset and evaluated the inference performance, communication cost and the resulting model sizes. We provided empirical evidence that our method, under all settings and years tested outperformed the FedAvg counterpart. Furthermore, in most settings, they demonstrated similar or even marginally superior performance compared to the centralized baseline. We have repeatedly seen in literature that moderately pruned models (before reaching extreme sparsities) tend to perform better than the unpruned counterpart \cite{https://doi.org/10.48550/arxiv.1506.02626, suzuki2018spectral}, and this coincides with our finding even in a distributed setting. However, we refrain from drawing any conclusive statement regarding this, as further empirical evaluation with different datasets and models would be required to evaluate this. Therefore, direct future work aims to explore this behaviour and the effects of different pruning policies with other open-source datasets and problem settings. Furthermore, model sparsification may be used in conjunction with other techniques and hardware to maximize the compression effects. Moving forward we aim to develop a pipeline to further improve on-device efficiency with the use of other methods.
	
	\section*{CRediT authorship contribution statement}
	
	\textbf{Andy Li:} Conceptualisation, Methodology, Software, Validation, Formal Analysis, Investigation, Data curation, Writing -- original draft preparation, Visualization. \textbf{Milan Markovic:} Validation, Investigation, Funding acquisition, Supervision, Writing -- review \& editing.  \textbf{Peter Edwards:} Resources, Funding acquisition, Supervision, Investigation,  Writing -- review \& editing.  \textbf{Georgios Leontidis:} Conceptualisation, Methodology, Investigation, Resources, Funding acquisition, Supervision, Writing -- review \& editing.
	
	\section*{Declaration of competing interest}
	
	The authors declare that they have no competing financial
	interests or personal relationships that could have appeared to influence the work reported in this paper.
	\section*{Data availability}
	The dataset used in this paper was downloaded from  \url{https://github.com/saeedkhaki92/CNN-RNN-Yield-Prediction}.
	\section*{Acknowledgements}
	
	The work described here was funded by the EPSRC 'Enhancing Agri-Food Transparent Sustainability' (EATS) project (grant number: EP/V042270/1) and by a University of Aberdeen PhD studentship.
	
	\bibliographystyle{elsarticle-num-names-alphsort}

	
		
		
		
		% \end{thebibliography}
\end{document}